\documentclass[10pt,twocolumn,letterpaper]{article}

\usepackage[pagenumbers]{cvpr} 

\newcommand{\cyp}[1]{\textcolor{black}{#1}}\newcommand{\whx}[1]{\textcolor{black}{#1}}\newcommand{\lyt}[1]{\textcolor{black}{#1}}

\definecolor{cvprblue}{rgb}{0.21,0.49,0.74}
\usepackage[pagebackref,breaklinks,colorlinks,allcolors=cvprblue]{hyperref}

\def\confName{CVPR}
\def\confYear{2026}

\title{\cyp{FACE: A Face-based Autoregressive Representation for High-Fidelity and Efficient Mesh Generation}}

\author{
    Hanxiao Wang$^{1,2}$ \quad 
    Yuan-Chen Guo$^{3}$ \quad 
    Ying-Tian Liu$^{3}$ \quad 
    Zi-Xin Zou$^{3}$ \\
    Biao Zhang$^{4}$ \quad 
    Weize Quan$^{1,2}$ \quad 
    Ding Liang$^{3}$ \quad 
    Yan-Pei Cao$^{3}$ \quad 
    Dong-Ming Yan$^{1,2}$ \\
    $^{1}$CASIA \quad $^{2}$UCAS \quad $^{3}$VAST \quad $^{4}$KAUST \\
}
\begin{document}

\twocolumn[{%
\renewcommand\twocolumn[1][]{#1}%
\maketitle
\includegraphics[width=\linewidth]{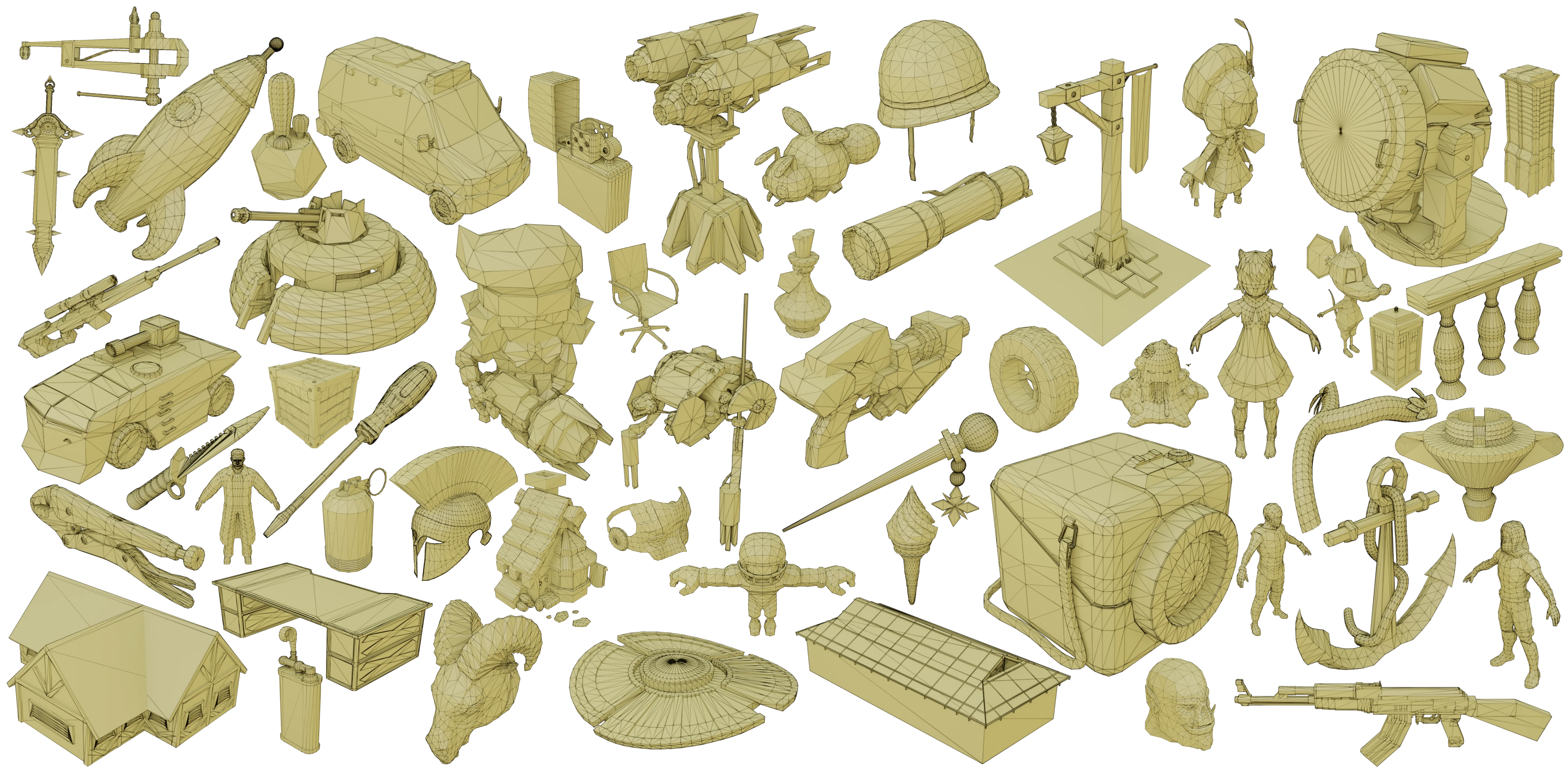}
\captionof{figure}{\textbf{High-fidelity meshes \whx{reconstructed} by FACE from point clouds.} We introduce \textbf{FACE}, a novel Autoregressive Autoencoder (ARAE) driven by a new mesh compression strategy. This paradigm represents the mesh using a dramatically shorter sequence, achieving state-of-the-art efficiency while producing high-quality 3D geometry. \vspace{1em}}

\label{fig:teaser}
}]
\begin{abstract}Autoregressive models for 3D mesh generation suffer from a fundamental limitation: they flatten meshes into long vertex-coordinate sequences. This results in prohibitive computational costs, hindering the efficient synthesis of high-fidelity geometry. We argue this bottleneck stems from operating at the wrong semantic level. We introduce \textbf{FACE}, a novel Autoregressive Autoencoder (ARAE) framework that reconceptualizes the task by generating meshes at the face level. Our ``\textbf{one-face-one-token}'' strategy treats each triangle face, the fundamental building block of a mesh, as a single, unified token. This simple yet powerful design reduces the sequence length by a factor of nine, leading to an unprecedented compression ratio of ~\textbf{0.11}, halving the previous state-of-the-art. This dramatic efficiency gain does not compromise quality; by pairing our face-level decoder with a powerful VecSet encoder, FACE achieves state-of-the-art reconstruction quality on standard benchmarks. The versatility of the learned latent space is further demonstrated by training a latent diffusion model that achieves high-fidelity, single-image-to-mesh generation. FACE provides a simple, scalable, and powerful paradigm that lowers the barrier to high-quality structured 3D content creation.\end{abstract}  
\section{Introduction}
\label{sec:intro}

\cyp{The generation of 3D content is a cornerstone of modern computing, fundamental to domains from industrial design and scientific visualization to virtual reality and gaming. Among various 3D representations, triangle meshes remain the industry standard, prized for their rendering efficiency and ability to represent complex topologies. Consequently, the direct generation of high-fidelity, topologically coherent meshes~\cite{pietroni2022hex} stands as a ``holy grail'' in computer graphics and vision. However, directly modeling the intricate structure of vertices and faces remains a significant and unsolved challenge.}

\cyp{Deep learning has driven substantial progress, with autoregressive (AR) models emerging as the dominant paradigm for end-to-end mesh generation~\cite{siddiqui2024meshgpt,chen2025meshxl,chen2024meshanything}. These methods flatten a mesh's structure into a one-dimensional sequence of vertex coordinates and generate it token by token. While this approach avoids the error accumulation of earlier two-stage methods~\cite{nash2020polygen}, it suffers from a fatal flaw: the crippling computational cost of the Transformer's self-attention mechanism. The resulting $\mathcal{O}(N^2)$ complexity with respect to sequence length creates a severe bottleneck, making the generation of high-resolution meshes with thousands of faces computationally prohibitive.}

\cyp{A flurry of research~\cite{chen2024meshanything1,weng2024scaling,lionar2025treemeshgpt,tang2024edgerunner,wang2025nautilus} has sought to mitigate this bottleneck through various compression strategies. One line of work focuses on complex traversal algorithms to optimize vertex reuse~\cite{chen2024meshanything1,tang2024edgerunner}, while another explores sophisticated tokenization schemes~\cite{wang2025nautilus,weng2024scaling} such as block indexing~\cite{weng2024scaling}. Although these methods have improved compression ratios, they often introduce new trade-offs: traversal-based strategies can be brittle and disrupt the mesh's global structure, while block-based tokenization can lead to an exploding vocabulary size. We argue that these approaches address the symptom, i.e., long sequence length, but not the underlying cause.}

\cyp{The fundamental issue lies in operating at the wrong semantic level. We introduce \textbf{FACE}, a novel \textbf{Autoregressive Autoencoder (ARAE)} framework that reconceptualizes the task by generating meshes \textit{at the level of faces}, not individual vertices. Our key insight is a \textbf{``one-face-one-token''} strategy: we treat each triangle face, the fundamental building block of a mesh, as a single token. This simple yet powerful design choice elevates the generation process to a higher semantic level, directly reducing the sequence length for the expensive attention mechanism by a factor of nine.}

\cyp{This conceptual shift yields a cascade of benefits. \textit{First}, it leads to an \textbf{unprecedented efficiency improvement}, achieving a state-of-the-art compression ratio of \textbf{0.11} and effectively halving the sequence length of the previous best method. \textit{Second}, this efficiency does not come at the cost of quality. By pairing our face-level decoder with a powerful VecSet~\cite{zhang20233dshape2vecset} encoder, our ARAE framework learns a robust latent space that achieves \textbf{state-of-the-art reconstruction quality} across standard benchmarks. \textit{Finally}, we demonstrate the \textbf{versatility of this learned latent space} by training a latent diffusion model that enables high-fidelity, single-image-to-mesh synthesis. This validates that FACE learns a truly meaningful and generalizable representation for 3D shapes.}

The main contributions of this work are as follows:

\begin{itemize}
\item We propose \textbf{FACE}, a novel framework that reconceptualizes mesh generation with a highly efficient ``one-face-one-token'' strategy.
\item We achieve a new \textbf{state-of-the-art compression ratio of 0.11}, doubling the efficiency of prior autoregressive models and fundamentally lowering the computational barrier for high-fidelity generation.
\item We demonstrate \textbf{state-of-the-art mesh reconstruction quality} on multiple challenging benchmarks, proving that our efficiency gains do not compromise fidelity.
\item We validate the quality of our learned latent space by enabling a downstream image-to-mesh generation task, showcasing its robustness and utility for multi-modal applications.
\end{itemize}
\section{Related Work}

\noindent\textbf{Mesh Reconstruction.}
Traditional mesh reconstruction from 3D data follows two main paradigms. Direct methods like the Ball-Pivoting~\cite{bernardini2002ball} and Voronoi-based approaches~\cite{amenta1998new} build connectivity from point sets but are often sensitive to noise. The more dominant paradigm extracts an isosurface from an implicit field, using classic algorithms like Marching Cubes~\cite{lorensen1998marching} and Dual Contouring~\cite{ju2002dual} that trade feature preservation for robustness. To address their limitations, learning-based successors like Neural Marching Cubes~\cite{chen2021neural} and Neural Dual Contouring~\cite{chen2022neural} leverage neural networks to reconstruct sharp features more reliably from varied inputs. However, all these methods are fundamentally for reconstruction, i.e., extracting a mesh from a given dense representation, rather than generation.

\noindent\textbf{3D Shape Representation and Generation.}
Generative 3D modeling~\cite{wu20153d,wu2016learning,xie2020generative,achlioptas2018learning,luo2021diffusion,chen2019learning,zhang20233dshape2vecset,triposg,zhang2024clay,zhao2025hunyuan3d} has explored a wide array of representations beyond meshes. Voxel-based methods~\cite{wu20153d,wu2016learning,xie2020generative} provide a direct volumetric grid but are limited in resolution by their cubic memory growth. Point clouds~\cite{achlioptas2018learning,luo2021diffusion,yang2019pointflow} offer flexibility but lack inherent surface topology. Recently, implicit neural representations like SDFs and Occupancy Networks~\cite{chen2019learning,zhang20233dshape2vecset,triposg,zhang2024clay,zhao2025hunyuan3d} have become dominant, encoding continuous geometry for memory-efficient, high-resolution modeling. Subsequent works like TRELLIS~\cite{xiang2025structured}, SparseFlex~\cite{he2025sparseflex} and OctGPT~\cite{wei2025octgpt} have further scaled these methods. However, a critical drawback unites these non-mesh paradigms: they all require a post-processing step like Marching Cubes~\cite{lorensen1998marching} to extract a final polygonal mesh. This indirect extraction offers no fine-grained control over the resulting face structure and can introduce artifacts, motivating methods for direct mesh generation.

\noindent\textbf{Direct Mesh Generation.}
Research on direct polygon mesh generation aims to construct 3D models with explicit topology in an end-to-end manner. Early two-stage methods like PolyGen~\cite{nash2020polygen} first predicted vertices and then their connectivity, a process prone to error accumulation. This was largely superseded by the end-to-end autoregressive paradigm pioneered by MeshGPT~\cite{siddiqui2024meshgpt} and adopted by others like MeshXL~\cite{chen2025meshxl} and PivotMesh~\cite{weng2024pivotmesh}, which flatten meshes into long 1D token sequences. This shift, however, introduced a new primary challenge: the prohibitive computational cost of these long sequences. Consequently, a significant body of research has focused on sequence compression. These efforts can be broadly categorized into three technical routes. \textbf{1) Traversal and Adjacency:} One line of work, including EdgeRunner~\cite{tang2024edgerunner}, MeshAnythingV2~\cite{chen2024meshanything1}, and MeshSilksong~\cite{song2025mesh}, optimizes vertex reuse by carefully ordering faces to compress the sequence. \textbf{2) Tokenization Strategies:} Another direction innovates on the token representation itself, with methods like TreeMeshGPT~\cite{lionar2025treemeshgpt} using a ``one-vertex-one-token'' approach, BPT~\cite{weng2024scaling} and Nautilus~\cite{wang2025nautilus} employing block indexing, and FreeMesh~\cite{liu2025freemesh} using coordinate merging. \textbf{3) Architectural Improvements:} A third route seeks efficiency via novel architectures, such as Meshtron's~\cite{hao2024meshtron} hierarchical structure and iFlame's~\cite{wang2025iflame} mixed-attention mechanisms. 
Recent works have further expanded the field by integrating Large Language Models~\cite{fang2025meshllm,wang2024llama} or reinforcement learning~\cite{zhao2025deepmesh,liu2025mesh}. While these compression efforts have advanced the state-of-the-art, they often introduce new trade-offs, from algorithmic complexity to exploding vocabulary sizes.
Notably, the idea of a ``face-per-token'' representation has been explored in diffusion-based models like PolyDiff~\cite{alliegro2023polydiff} and MeshCraft~\cite{he2025meshcraft}. However, the non-sequential nature of diffusion makes it challenging to generate variable-length outputs and ensure topological completeness. To our knowledge, successfully implementing a face-as-token paradigm within an \textbf{autoregressive} framework, which naturally handles variable sequence lengths and sequential dependencies, has not been previously achieved. Our work is the first to demonstrate this, unlocking the efficiency of a face-level representation within a flexible and powerful sequential generation process.

\begin{figure}[!t] 
    \centering
    \includegraphics[width=1.0\linewidth]{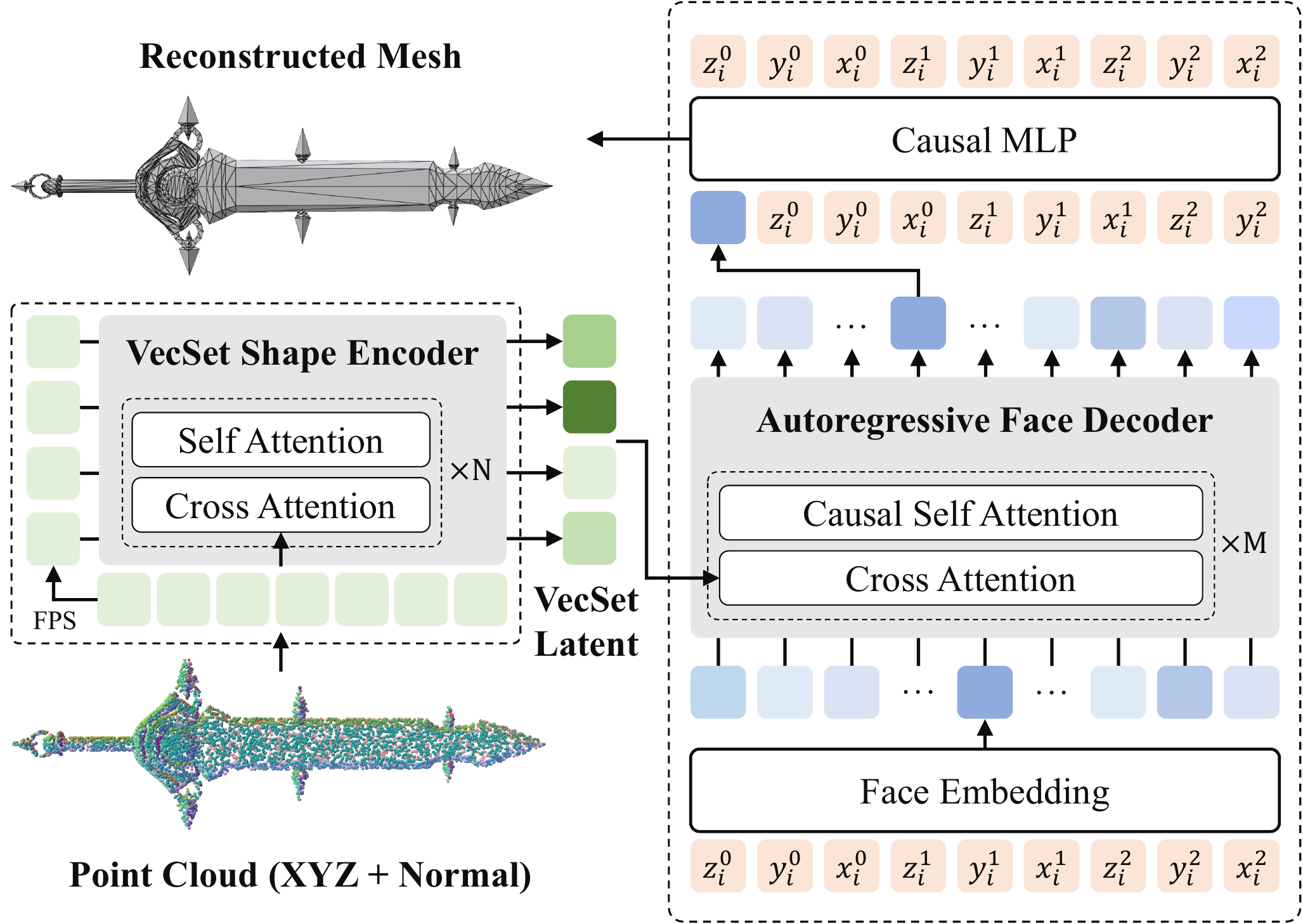}
    \caption{The end-to-end pipeline of our \textbf{FACE} model. An encoder compresses the input point cloud into a latent VecSet. An autoregressive decoder then conditions on this VecSet, generating the mesh face-by-face. A Face Embedding layer encoder 9 tokens of each face and a CausalMLP head decodes each latent face token into 9 quantized coordinate tokens.}

    \label{fig:teaser_full}
\vspace{-6mm}
\end{figure}

\section{Methodology}
\label{sec:method}

Our methodology centers on \textbf{FACE}, \cyp{a novel framework designed to model the conditional probability $p(M|P)$, where $M$ is a high-fidelity triangle mesh and $P$ is its corresponding input shape (in the point cloud representation). We formulate this as an \textbf{Autoregressive Autoencoder (ARAE)}~\cite{tang2024edgerunner}, comprising two main components: (1) a \textbf{Shape Encoder} that learns to compress the input $P$ into a compact latent representation $C$, i.e., a VecSet~\cite{zhang20233dshape2vecset} and (2) an \textbf{Autoregressive Decoder} that reconstructs the mesh $M$ by generating its constituent faces sequentially, conditioned on $C$.}

\cyp{While powerful, existing autoregressive approaches are severely hampered by the quadratic computational complexity of the Transformer's self-attention mechanism when applied to the long token sequences derived from flattened meshes. FACE overcomes this fundamental bottleneck by reconceptualizing the generation process. Instead of operating on low-level vertex coordinates, our framework operates at the higher semantic level of triangle faces. This is achieved through a novel ``\textbf{one-face-one-token}'' strategy, which drastically shortens the sequence length and unlocks scalable, high-fidelity generation. The end-to-end pipeline of our ARAE framework is illustrated in~\cref{fig:teaser_full}.}

\subsection{\cyp{Face-based Autoregressive Representation}}
\cyp{To model the mesh $M$ autoregressively, we first represent it as a deterministic, ordered sequence of its constituent faces, $F = (f_1, f_2, \ldots, f_N)$, where $N$ is the total number of faces.}

\noindent \textbf{Face Ordering.} 
\cyp{Establishing a consistent and effective ordering is critical for autoregressive modeling. While prior works have explored complex graph traversal algorithms~\cite{weng2024scaling,lionar2025treemeshgpt,chen2024meshanything1} to optimize for vertex locality, we find that a surprisingly simple and deterministic spatial ordering is highly effective for face-level generation. Specifically, we sort all faces $f_i$ based on the lexicographical $\text{ZYX}$ order~\cite{siddiqui2024meshgpt,chen2025meshxl} of their minimum-coordinate vertex. This approach provides a canonical ordering for any given mesh, eliminating a major source of system complexity without compromising generation quality, as our experiments in~\cref{sec:ablation} demonstrate.}

\noindent \textbf{The ``One-Face-One-Token'' Strategy.}
\cyp{The cornerstone of our framework is the ``one-face-one-token'' strategy. Each face $f_i$ in the sequence is defined by the coordinates of its three vertices, flattened into a 9-dimensional vector: $f_i = (v^0_i, v^1_i, v^2_i) \in \mathbb{R}^9$. Instead of further decomposing each face into nine separate coordinate tokens (the standard approach which results in a sequence of length $9D$), we treat the entire 9D face vector $f_i$ as a single unit. As detailed in~\cref{sec:architecture}, a lightweight embedding layer projects this 9D vector into a single $d_{\text{model}}$-dimensional token that is fed into the Transformer decoder. This design choice directly attacks the $\mathcal{O}(S^2)$ computational bottleneck by reducing the sequence length for the self-attention mechanism by a factor of nine. This allows FACE to model significantly more complex meshes within a given computational budget, paving the way for truly scalable generation.}

\subsection{\cyp{Model Architecture}}
\label{sec:architecture}
\cyp{Our ARAE framework is realized through a carefully designed encoder-decoder architecture. The encoder compresses the input point cloud into a compact and powerful latent representation, while the decoder leverages this representation to autoregressively synthesize the mesh, face by face.}

\subsubsection{{\cyp{Shape Encoder}}}
\cyp{The primary role of the Shape Encoder is to map the raw input point cloud $P \in \mathbb{R}^{m \times 3}$ into a compact latent VecSet $C \in \mathbb{R}^{k \times d_{\text{latent}}}$ that captures the global geometry of the shape. To achieve this, we adopt the powerful and proven architecture from 3DShape2VecSet~\cite{zhang20233dshape2vecset}, which excels at learning robust 3D representations.}

\cyp{The encoding process begins by using Farthest Point Sampling (FPS)~\cite{moenning2003fast} to select a smaller, representative set of $k$ query points from the input. These queries then attend to the full point set via a cross-attention mechanism, aggregating global geometric information into an initial set of $k$ vectors, $C'$:}
\begin{equation}
C' = \text{CrossAttn}(Q=Q, K=K_P, V=V_P)
\end{equation}
\cyp{where $Q$, $K_P$, and $V_P$ are projections of the query points and the full point set, respectively. This initial representation $C'$ is then refined by a standard Transformer Encoder stack of $L_E$ layers, producing the final latent VecSet $C$:}

\begin{equation}
C = \text{TransformerEncoder}_{L_E}(C').
\end{equation}
\cyp{This final VecSet $C$ serve as the comprehensive global conditioning signal for the decoder, providing the necessary context to guide the face generation process.}

\subsubsection{Autoregressive Face Decoder}
\cyp{The Autoregressive Face Decoder is tasked with generating the mesh's face sequence $F = (f_1, f_2, \ldots, f_N)$ conditioned on the latent VecSet $C$. Crucially, unlike methods that rely on a separately trained VAE for coordinate tokenization (e.g., MeshGPT~\cite{siddiqui2024meshgpt}), \textit{our decoder learns its own internal representation for faces as part of a single, end-to-end training loop}. This allows the representation to be optimized directly for the generation task.}

\cyp{The generation process for the $i$-th face $f_{i}$ proceeds through a sequence of carefully defined steps within a Transformer architecture of $L_D$ layers:}

\noindent\cyp{\textbf{\textit{1. Face Embedding.}} At each step $i$, the ground-truth faces from the prefix serves as the input. Each face $f_{i-1} \in \mathbb{R}^9$ is first projected into a $d_{\text{model}}$-dimensional token $t_{i-1}$ by a lightweight MLP, which we term the \textbf{Face Pooling} layer:}
\begin{equation}
    t_{i-1} = \text{MLP}_{\text{embed}}(f_{i-1}).
\end{equation}
\cyp{This layer is the architectural realization of our ``one-face-one-token'' strategy, mapping the continuous coordinate space of a face into a single latent token for the Transformer.}

\noindent\cyp{\textbf{\textit{2. Transformer Processing.}} The sequence of face tokens $(t_1, t_2, \ldots, t_{i-1})$ is then processed by the Transformer decoder. Each layer $l$ in the decoder performs two key attention operations:}
\begin{align}
    H'_{l} &= \text{CausalSelfAttn}(H_l) \\
    H_{l+1} &= \text{CrossAttn}(Q=H'_{l}, K=C, V=C)
\end{align}
\cyp{First, \textbf{Causal Self-Attention} allows the model to look at the sequence of previously generated face tokens $t_{<i}$, capturing the local structure and connectivity of the mesh. Second, \textbf{Cross-Attention} takes the output of the self-attention as its query and uses the encoder's VecSet $C$ as its key and value. This step injects the global shape context at every layer, ensuring that each local generation decision is informed by the overall target geometry. The final output of the Transformer stack at step $i$ is a latent face vector $h_i \in \mathbb{R}^d_{\text{model}}$.}

\begin{table*}[!t]
  \centering
  \caption{Comparison on mesh token efficiency. Our method achieves a state-of-the-art compression ratio, representing the mesh with the shortest sequence length.}
  \label{tab:tokenization_efficiency_full}
  \resizebox{\textwidth}{!}{%
    \begin{tabular}{lcccccccccccc}
      \toprule
      Method & MeshXL & MeshAnything & MeshGPT & PivotMesh & EdgeRunner &  MeshAnything v2 & DeepMesh & Nautilus & BPT & Mesh-Silksong & TreeMeshGPT & \textbf{Ours} \\
      \midrule
      Compression Ratio $\downarrow$ & 1.00 & 1.00 & 0.67 & 0.67 &  0.47 & 0.46 & 0.28 & 0.27 & 0.26 & 0.22 & 0.22 & \textbf{0.11} \\
      \bottomrule
    \end{tabular}%
  }
  \vspace{-2mm}
\end{table*}

\noindent\cyp{\textbf{\textit{3. Face Decoding Head.}} The final step is to decode the latent vector $h_i$ back into the nine quantized coordinates of the face $f_i$. For this, we employ a specialized CausalMLP~\cite{lionar2025treemeshgpt} head. This lightweight module introduces a second level of autoregression within the face itself. The prediction for the $j$-th coordinate token of the face is conditioned not only on the latent vector $h_i$ but also on all previously predicted coordinate tokens within that same face. This enforces a causal dependency among the coordinate predictions, which we find is significantly more effective than attempting to predict all nine coordinates in parallel.}

This \textit{hierarchical} process, i.e., autoregressive at the face-level via the Transformer and at the coordinate-level via the CausalMLP, allows FACE to robustly generate complex mesh structures in a coherent, end-to-end manner.

\subsection{Training Objective and Efficiency Analysis}
\label{sec:token_efficiency}

\subsubsection{End-to-End Training Objective}
\cyp{The entire FACE framework is trained end-to-end by minimizing a single, unified objective: the reconstruction loss for the mesh faces. For each face $f_i$ in the ground-truth sequence, our CausalMLP head predicts a sequence of logit vectors $(L_{i,1}, ..., L_{i,9})$, one for each of the nine quantized coordinates. The training objective is to minimize the sum of the standard cross-entropy losses for each coordinate prediction, averaged over all faces in the mesh.}

Let $c_{i,j}$ be the ground-truth token index for the $j$-th coordinate of the $i$-th face. The total loss $\mathcal{L}$ is defined as:
\begin{equation}
    \mathcal{L} = \frac{1}{N} \sum_{i=1}^N \sum_{j=1}^9 \text{CrossEntropy}(L_{i,j}, c_{i,j}).
\end{equation}
\cyp{By optimizing this objective, the encoder and decoder jointly learn to find a compact latent representation $C$ from which the original mesh $M$ can be faithfully reconstructed. This end-to-end formulation ensures that the learned representations are directly optimized for high-fidelity generative modeling.}

\subsubsection{Efficiency Analysis}
The architectural design of FACE translates directly into a state-of-the-art leap in tokenization efficiency. This improvement is a direct consequence of our ``one-face-one-token'' strategy.

The primary computational bottleneck in autoregressive models is the self-attention mechanism, whose complexity is quadratic, $\mathcal{O}(S^2)$, with respect to the sequence length $S$. Previous methods~\cite{chen2024meshanything,chen2025meshxl}, which tokenize each of a face's nine coordinates, must process a sequence of length $S = 9 \times |F|$. Our approach, by contrast, operates on a sequence of length $S = |F|$. This $9\times$ reduction in sequence length leads to a theoretical $81\times$ \lyt{computational cost reduction} and \lyt{roughly $9\times$} memory footprint \lyt{reduction} of the self-attention layers \lyt{with flash attention~\cite{dao2022flashattention,dao2023flashattention}}.
As shown in~\cref{tab:tokenization_efficiency_full}, we define the practical Compression Ratio as the sequence length a model must process relative to the baseline of one token per coordinate. FACE achieves a compression ratio of $0.11$, a $2\times$ improvement over the previous best of $0.22$. 

Crucially, this efficiency is achieved through architectural elegance, not complex, lossy compression schemes. The Face Pooling and CausalMLP modules add only a negligible linear, $\mathcal{O}(N)$, computational cost. 
This allows FACE to scale to higher resolutions and more complex geometries, fundamentally advancing the frontier of what is possible in direct mesh generation.

\subsection{Image-to-Mesh via Latent Diffusion}
\label{sec:img2mesh}
\begin{figure}[!h]
    \centering
    \includegraphics[width=1.0\linewidth]{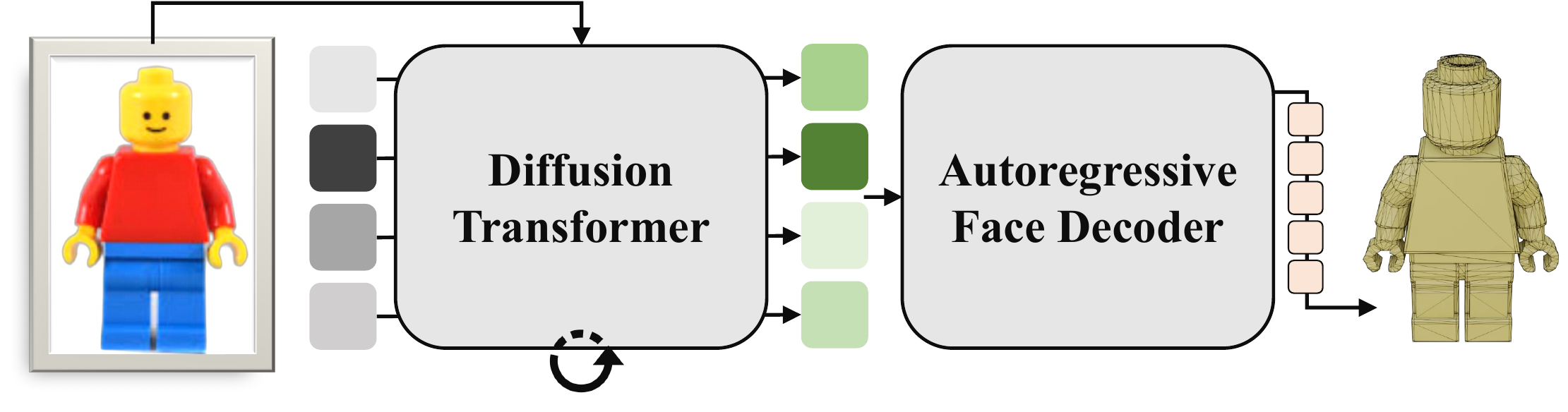}
     \vspace{-15pt}
    \caption{Overview of our image-to-mesh generation pipeline. We first use the input image to condition a DiT model. The resulting latent VecSet is then fed into the Autoregressive Face Decoder to produce the final mesh.}
    \label{fig:diffusion}
    \vspace{-5mm}
\end{figure}

A central hypothesis of our ARAE formulation is that by training the model end-to-end, the shape encoder learns a rich, structured, and semantically meaningful latent space $C$ for 3D meshes. To rigorously validate this hypothesis and demonstrate the utility of our learned representations, we apply them to single-image-to-mesh generation which is a challenging downstream task.

As illustrated in \cref{fig:diffusion}, our image-to-mesh pipeline consists of two stages.

First we train a Diffusion Transformer (DiT)~\cite{peebles2023scalable} to perform denoising process within the latent space, conditioning on the image features extracted by a pretrained DINOv3 model~\cite{simeoni2025dinov3}. 

Then the pretrained FACE decoder translates the generated latent VecSet into high-fidelity meshes face by face.

The success of this pipeline hinges on the quality and robustness of the latent space learned by the FACE autoencoder. The ability to produce high-quality meshes from image-conditioned VecSets (without any fine-tuning of the FACE decoder) provides powerful evidence that our ARAE framework has successfully captured a versatile and fundamental representation of 3D shape. This demonstrates not only the effectiveness of our core method but also its potential as a foundational component for future multi-modal generative 3D workflows.
\begin{table*}[!t]
 \centering
 \caption{Quantitative comparison of mesh reconstruction quality on multiple datasets.}
 \vspace{-1mm}
 \label{tab:combined_results_wide}
 \small
 \begin{tabular}{lcccccc}
 \toprule
& \multicolumn{2}{c}{\textbf{Objaverse}} & \multicolumn{2}{c}{\textbf{Toys4K}} & \multicolumn{2}{c}{\textbf{Famous}} \\
 \cmidrule(lr){2-3} \cmidrule(lr){4-5} \cmidrule(lr){6-7}
 Method & Hausdorff $\downarrow$ & Chamfer $\downarrow$ & Hausdorff $\downarrow$ & Chamfer $\downarrow$ & Hausdorff $\downarrow$ & Chamfer $\downarrow$ \\
 \midrule
 MeshAnything & 0.327 & 0.128 & 0.264 & 0.110 & 0.399 & 0.177 \\
 MeshAnythingV2 & 0.265 & 0.089 & 0.238 & 0.090 & 0.488 & 0.203 \\
 TreeMeshGPT & 0.183 & 0.055 & 0.133 & 0.047 & 0.226 & 0.064 \\
 BPT & 0.126 & 0.043 & 0.091 & 0.037 & 0.143 & 0.061 \\
 \midrule
 \textbf{Ours (FACE)} & \textbf{0.090} & \textbf{0.041} & \textbf{0.067} & \textbf{0.033} & \textbf{0.077} & \textbf{0.049} \\
 \bottomrule
 \end{tabular}
 \vspace{-3mm}
\end{table*}

\section{Experiment}
\label{sec:experiment}

\subsection{Implementation Details}

\paragraph{Autoregressive Autoencoder.}

We train an ARAE with a total of 500M parameters. As the decoder is directly responsible for accurately reconstructing the faces, we apply an asymmetric encoder-decoder design~\cite{triposg,gigatok} with the decoder being larger than the encoder. \lyt{The encoder consists of 8 layers with hidden dimension 768, and the encoder consists of 24 layers with hidden dimension 1024.} \lyt{We sample 8192 points with normals on the mesh surface as the encoder input. The VecSet latent is of 2048 tokens with bottleneck dimension 64. For training, we select the meshes with fewer than 4,000 faces from the Objaverse dataset~\cite{deitke2023objaverse} to form a subset with around 130,000 meshes.} The vertex positions of the mesh are normalized and quantized to integers in range $[0, 127]$. During training, we apply random rotation and flipping to the mesh, as well as random \lyt{individual} scaling along each axis. The model is trained with the Muon optimizer~\cite{liu2025muon,jordan2024muon} with a learning rate of $6 \times 10^{-4}$ and a weight decay of 0.1, for 100K steps on 8 NVIDIA A100 80GB GPUs. During inference, we use a deterministic sampling strategy that takes \lyt{the token with top-1 probability} autoregressively.

\paragraph{Image-Conditioned Diffusion Transformer.}

For image-conditioned generation, we train a diffusion transformer model (DiT)~\cite{rombach2022high} with 350M parameters using the standard flow matching~\cite{lipman2022flow,liu2022flow} objective. \lyt{The training dataset is a curated subset of 50,000 meshes from the ARAE training dataset. Each model is paired with 10 renderings with random lighting conditions and camera poses.} The DiT is trained for 400K steps on 32 NVIDIA A100 80GB GPUs. We use the Muon optimizer with a learning rate of $1 \times 10^{-4}$ without weight decay. During inference, we sample the latent for 100 steps using the Euler solver.

\subsection{Evaluation Settings}
\paragraph{Datasets.}
To evaluate the performance and generalization of our method, we test on three distinct datasets that were not involved during training:
\begin{itemize}
    \item \textbf{Objaverse}: A set of 500 randomly selected models from our Objaverse~\cite{deitke2023objaverse} test split.
    \item \textbf{Toys4K}: A collection of 900 models with fewer than 4000 faces from the Toys4K dataset~\cite{stojanov2021using}. 
    \item \textbf{Famous}: A set of 100 models widely used in 3D graphics research (e.g., elk, Armadillo), used to evaluate the generalization capability on complex and iconic shapes.
\end{itemize}

\noindent \textbf{Baselines.}
We compare the mesh reconstruction quality of our method (FACE) against several previous autoregressive mesh reconstruction methods: MeshAnything~\cite{chen2024meshanything}, MeshAnythingV2~\cite{chen2024meshanything1}, BPT~\cite{weng2024scaling}, and TreeMeshGPT~\cite{lionar2025treemeshgpt} \lyt{in~\cref{sec:mesh-reconstruction}}. 
\lyt{For image-conditioned generation, because the model weight is not released, we only compare with the official results from the EdgeRunner~\cite{tang2024edgerunner} project page in~\cref{sec:image-cond-gen}.
We also present the ablation results on key design choices of  FACE in ~\cref{sec:ablation}. Furthermore, we show the potential for FACE to achieve even better reconstruction quality by scaling up the model size and quantization resolution in ~\cref{sec:scaling}.}

\begin{figure*}[!h] 
    \centering
    \includegraphics[width=0.8\textwidth]{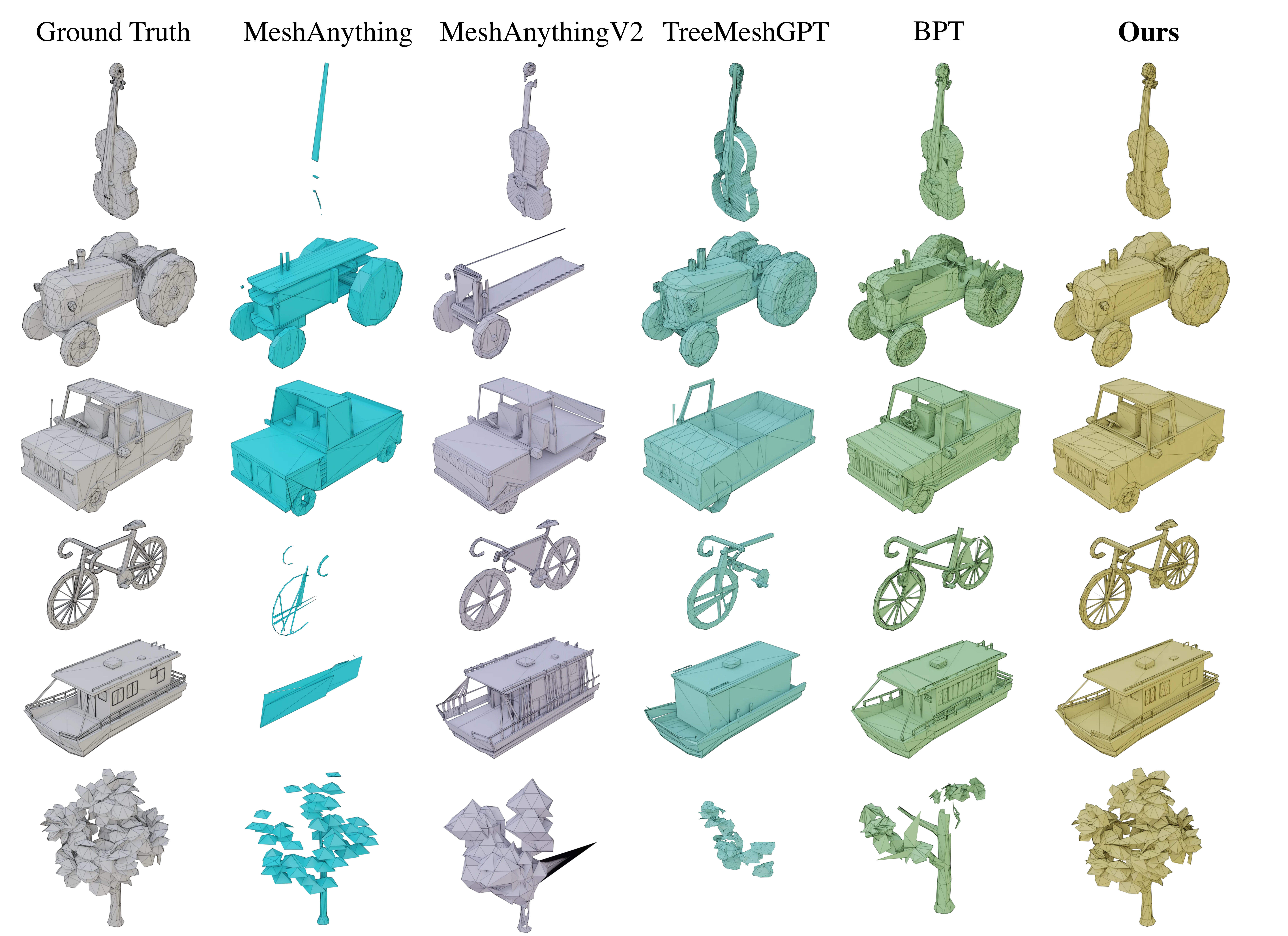}
   \vspace{-3mm}
    \caption{Qualitative comparison of mesh reconstruction quality on the \textbf{Toys4K} dataset.}
    \label{fig:qualitative_toy4k}
      \vspace{-3mm}
\end{figure*}

\subsection{\lyt{Mesh Reconstruction}}
\label{sec:mesh-reconstruction}

We evaluate the mesh reconstruction quality using the standard Hausdorff Distance (HD) and Chamfer Distance (CD) metrics. HD measures the ``worst-case'' deviation between models, while CD measures the ``average'' proximity between them.
As shown in~\cref{tab:combined_results_wide}, our method significantly outperforms all baselines across all test datasets.

On the large-scale \textbf{Objaverse} and \textbf{Toys4K} benchmarks, our method achieves the lowest error in both HD and CD, demonstrating superior reconstruction fidelity. \lyt{For instance, our method achieves over 26\% lower error than the best baseline method measured by Hausdorff Distance (0.067 v.s. 0.091).}

The results on the \textbf{Famous} dataset are particularly indicative of our model's strong generalization capabilities. These models are often complex and stylistically different from the training data. Our method again achieves state-of-the-art results, proving its robustness.

In~\cref{fig:qualitative_toy4k}, we provide a qualitative comparison against baseline methods on the Toys4K dataset. Meshes reconstructed by our method exhibit consistently higher fidelity and fewer topological errors. While baseline methods often produce artifacts such as unwanted holes, \lyt{incomplete components}, or overly smoothed-out details, our method produces clean surfaces that faithfully capture sharp features and intricate details from the input. The robust performance holds true even for complex models with numerous connected components, such as the tree in the final row, where our method produces the most complete and structurally sound result. 
\subsection{Image Conditioned Generation}
\label{sec:image-cond-gen}
\begin{figure*}[!t]
    \centering
    \includegraphics[width=0.8\textwidth]{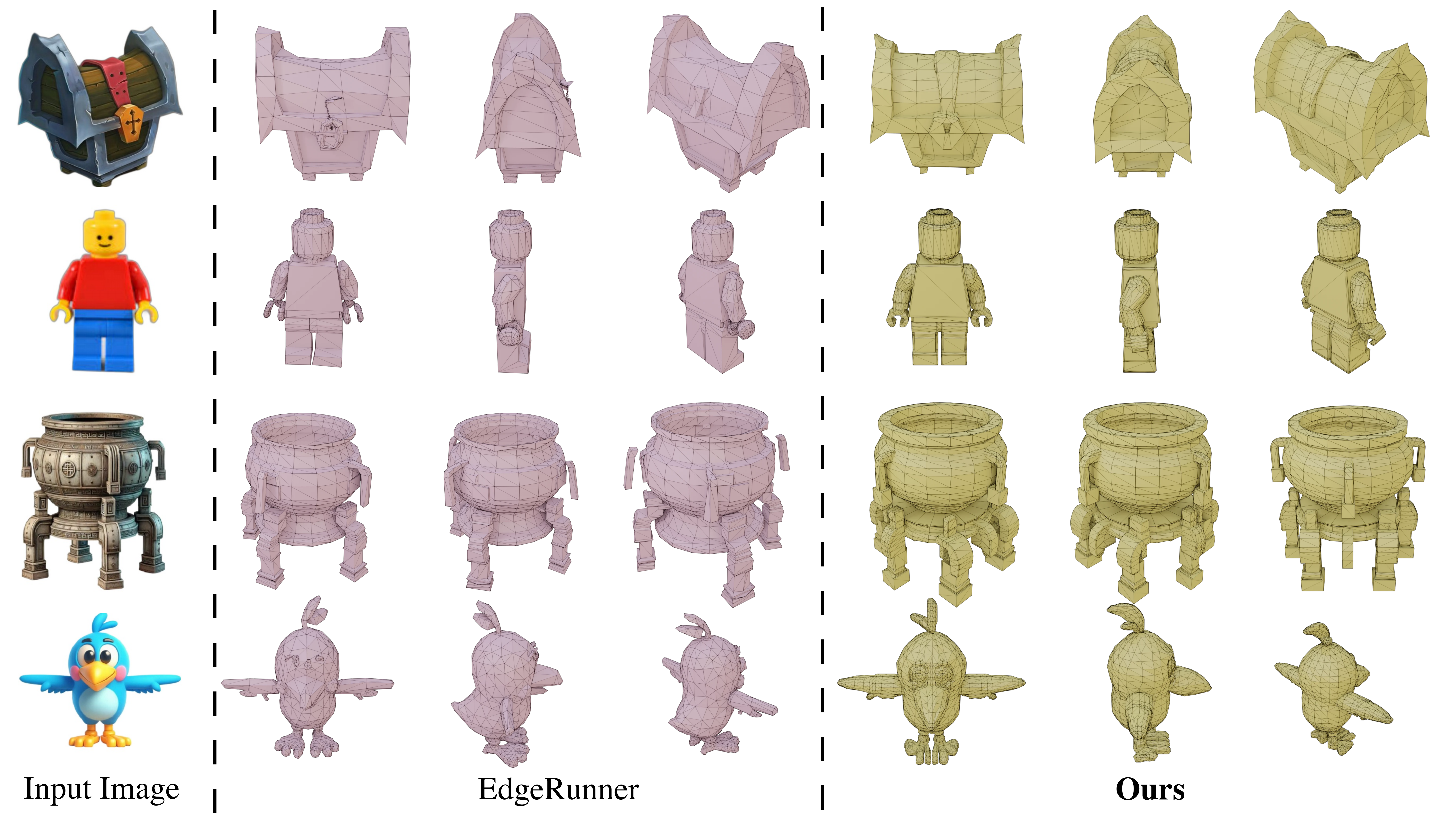}
    \vspace{-3mm}
    \caption{Qualitative comparison of image-conditioned mesh generation.}
    \label{fig:image-condition}
    \vspace{-3mm}
\end{figure*}

\lyt{Since previous image-conditioned mesh generation methods do not release their model,} we can only compare our results against the qualitative results for EdgeRunner~\cite{tang2024edgerunner} provided on their project page. As shown in~\cref{fig:image-condition}, our method achieves superior detail with greater alignment to the input image, along with improved topology connectivity. In contrast, Edgerunner often generates disconnected mesh fragments and incomplete structures. For example, our method successfully reconstructs fine-grained details like the hand and shoulder of the Lego figure (second row) and the distinct eye geometry of the bird (fourth row), where the baseline method fails.

\subsection{Ablation Studies}
\label{sec:ablation}
We conduct a series of ablation studies to validate several key design choices of FACE, namely mesh face ordering, the \lyt{query of} VecSet encoder, and the coordinate decoding head.
The experiments in this section utilize a decoder with $d_{\text{latent}}=768$.

\begin{table}[htbp]
  \centering
  \caption{Ablation on mesh face ordering strategies.}
   \vspace{-3mm}
  \label{tab:order_comparison}
  \small
  \scalebox{0.95}{
  \begin{tabular}{lcc}
    \toprule
    Method (Order) & Hausdorff Distance $\downarrow$ & Chamfer Distance $\downarrow$ \\
    \midrule
    BFS & 0.728 & 0.528 \\
    DFS & 0.171 & 0.077 \\
    ZYX-component & 0.110 & \textbf{0.045} \\
    \midrule 
    \textbf{ZYX} & \textbf{0.103} & 0.047 \\
    \bottomrule
  \end{tabular}}
  \vspace{-3mm}
\end{table}

\begin{table}[htbp]
  \centering
  \caption{Ablation on different queries of the shape encoder.}
  \vspace{-3mm}
  \label{tab:encoder_ablation}
  \small
  \begin{tabular}{lcc}
    \toprule
    Queries & Hausdorff Distance $\downarrow$ & Chamfer Distance $\downarrow$ \\
    \midrule
    Learnable & 0.132 & 0.058 \\
    \textbf{Downsample} & \textbf{0.103} & \textbf{0.047} \\
    \bottomrule
  \end{tabular}
  \vspace{-3mm}
\end{table}

\begin{table}[htbp]
  \centering
  \caption{Ablation on coordinate decoding strategies.}
  \vspace{-3mm}
  \label{tab:decoder_ablation}
  \small
  \begin{tabular}{lcc}
    \toprule
    Method & Hausdorff Distance $\downarrow$ & Chamfer Distance $\downarrow$ \\
    \midrule
    Parallel Decode & 0.426 & 0.239 \\
    Attention-based & 0.132 & 0.064 \\
    \midrule 
    \textbf{CausalMLP} & \textbf{0.103} & \textbf{0.047} \\
    \bottomrule
  \end{tabular}
  \vspace{-3mm}
\end{table}

\noindent \textbf{Mesh Face Ordering.}
The order in which faces are generated is critical. We test four ordering strategies: \textbf{DFS} (Depth-First Search), \textbf{BFS} (Breadth-First Search) (similar to~\cite{lionar2025treemeshgpt}), \textbf{ZYX} (lexicographical, similar to~\cite{siddiqui2024meshgpt}), and \textbf{ZYX-component} (grouping by connected components first, then \lyt{following ZYX in each component}).
As shown in~\cref{tab:order_comparison}, spatial sorting (ZYX and ZYX-component) massively outperforms graph-traversal orders (DFS, BFS). The ZYX and ZYX-component strategies perform comparably, with ZYX achieving a slightly better Hausdorff Distance. We thus use the simpler ZYX ordering for all our experiments.

\noindent \textbf{Query of VecSet Encoder.}
We conduct an ablation study to validate our choice of the query tokens in VecSet encoder. We train two versions of our 128-resolution model: \lyt{one equipped with learnable queries and the other with the downsampled point cloud as the queries.} As shown in~\cref{tab:encoder_ablation}, \lyt{the one with downsampled queries achieves obviously better reconstruction accuracy, which is} consistent with the findings in 3DShape2VecSet~\cite{zhang20233dshape2vecset}.

\noindent \textbf{Coordinate Decoding Strategy.}
We test three alternative methods for decoding the latent face token $f_{\text{out}, i}$ into 9 coordinate tokens, as shown in~\cref{tab:decoder_ablation}. \lyt{The variants include} 
(1) \textbf{Parallel Decode}: A single MLP that outputs a $9 \times |V|$ tensor, which performs poorly. 
(2) \textbf{Attention-based}: An attention layer to decode the coordinates, which is less effective. 
(3) \textbf{CausalMLP}: The lightweight projection head.
The results clearly show that CausalMLP is the most effective one, achieving the best performance by a large margin.

\subsection{Scaling Up}
\label{sec:scaling}
To demonstrate the scalability of FACE, we train a larger-scale ARAE with 1.2B parameters (denoted as \textbf{Ours}-\textit{large}). The number of sampled input points is increased to 65,536, and the vertex positions are quantized to integers in range $[0, 1023]$, preserving intricate geometric details and reducing quantization errors. The model is trained on an internal dataset with 380K high-quality meshes.
\begin{figure}[!h]
    \centering
    \includegraphics[width=0.45\textwidth]{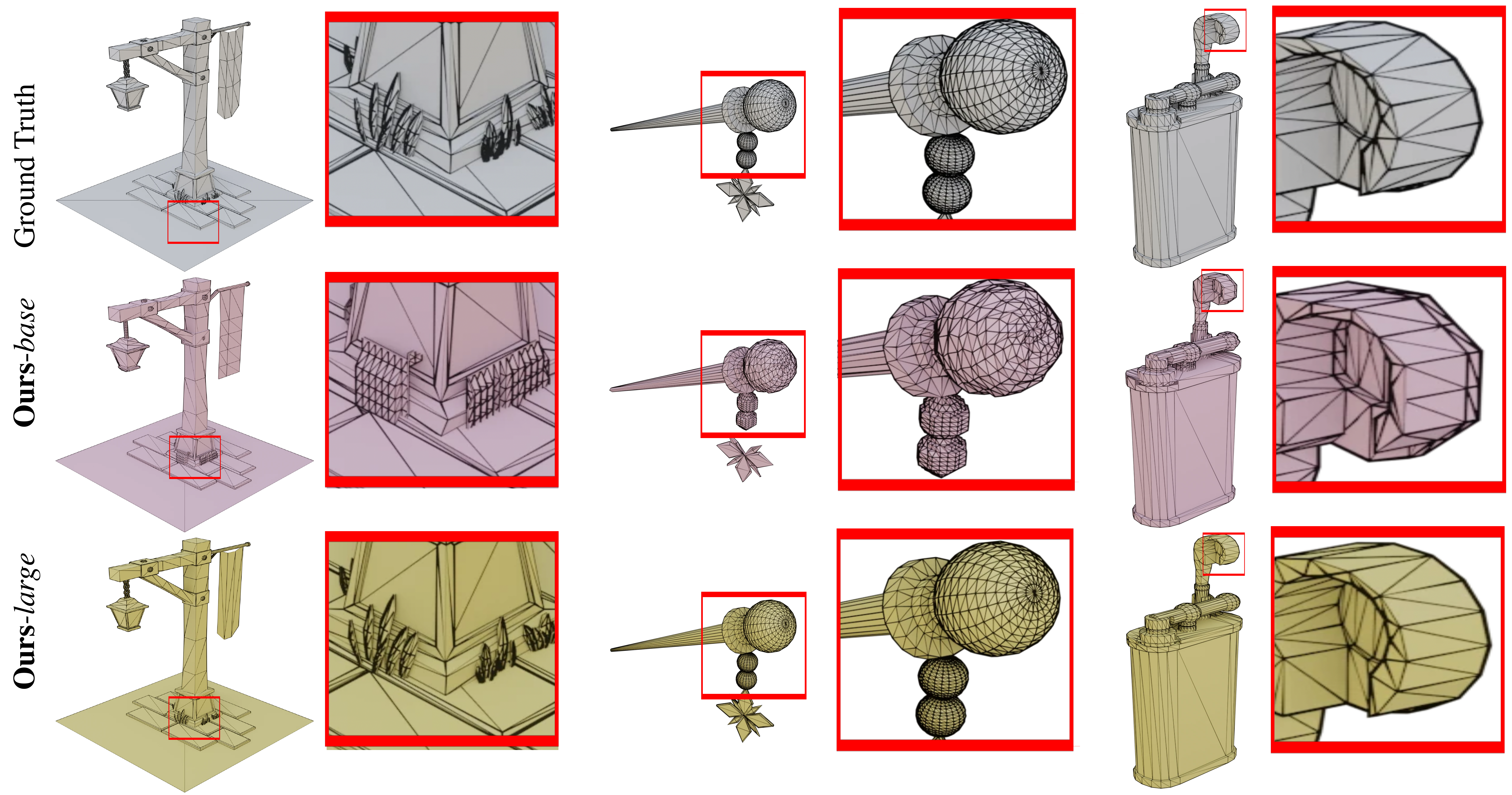}
    \vspace{-2mm}
    \caption{Qualitative comparisons between the base and large  model. }
    \label{fig:128vs1024}
     \vspace{-3mm}
\end{figure}

As shown in~\cref{fig:128vs1024}, our FACE architecture can reconstruct the target mesh with greater precision when scaling up in multiple aspects. Our large model not only achieves exceptional fidelity in overall shape but also excels at preserving fine-grained geometric details and sharp features. This shows evidence that the FACE framework possesses promising scaling properties, laying a solid foundation for high-fidelity 3D mesh representation.

\section{Conclusion}

In this work, we introduced \textbf{FACE}, a novel Autoregressive Autoencoder framework for autoregressive mesh generation, achieving state-of-the-art results in mesh reconstruction across standard datasets while simultaneously attaining a SOTA compression ratio of \textbf{0.11}.
We demonstrated the quality and versatility of the latent space learned by our method by training a latent diffusion model, which achieves high-fidelity, single-image-to-mesh generation. Furthermore, we confirmed the scalability of our approach by successfully training a high-resolution model.
We believe FACE offers a simple, powerful, and scalable paradigm for 3D generative modeling and provides a strong foundation for future research.

\noindent \textbf{Limitations.}
While we demonstrate scalability up to a 1024-resolution, this discrete representation is not infinite and still imposes an upper bound on the achievable level of detail. Second, our reliance on input point clouds means that extremely fine-grained or thin structures, such as the spokes of a bicycle wheel, may be inadequately sampled, potentially leading to incomplete reconstructions in these challenging areas.

{
    \small
    \bibliographystyle{ieeenat_fullname}
    \bibliography{main}
}


\end{document}